\documentclass[final,5p,times,twocolumn,authoryear]{elsarticle}

\usepackage{amssymb}
\usepackage{amsthm}
\usepackage{booktabs}
\usepackage{multirow}
\usepackage{lineno}

\journal{Computerized Medical Imaging and Graphics}

\begin{document}

\begin{frontmatter}

\title{Analysis of Vision-based Abnormal Red Blood Cell Classification}

\author[label1]{Annika Wong}
\author[label1]{Nantheera Anantrasirichai}
\author[label2]{Thanarat H. Chalidabhongse}
\author[label3]{Duangdao Palasuwan}
\author[label4]{Attakorn Palasuwan}
\author[label1]{David Bull}

\affiliation[label1]{organization={Visual Information Laboratory, University of Bristol},
              city={Bristol},
             postcode={BS8 1UB},
             country={UK}}
             
\affiliation[label2]{organization={Department of Computer Engineering, Chulalongkorn University},
              city={Bangkok},
             country={Thailand}}
             
\affiliation[label3]{organization={Research Group on Applied Computer Engineering Technology for Medicine and Healthcare, Chulalongkorn University},
              city={Bangkok},
             country={Thailand}}
             
\affiliation[label4]{organization={Cell Disorders Research Unit, Department of Clinical Microscopy, Faculty of Allied Health Sciences, Chulalongkorn University},
              city={Bangkok},
             country={Thailand}}

\begin{abstract}
Identification of abnormalities in red blood cells (RBC) is key to diagnosing a range of medical conditions from anaemia to liver disease. Currently this is done manually, a time-consuming and subjective process. This paper presents an automated process utilising the advantages of machine learning to increase capacity and standardisation of cell abnormality detection, and its performance is analysed. Three different machine learning technologies were used: a Support Vector Machine (SVM), a classical machine learning technology; TabNet, a deep learning architecture for tabular data; U-Net, a semantic segmentation network designed for medical image segmentation. A critical issue was the highly imbalanced nature of the dataset which impacts the efficacy of machine learning. To address this, synthesising minority class samples in feature space was investigated via  Synthetic Minority Over-sampling Technique (SMOTE) and cost-sensitive learning. A combination of these two methods is investigated to improve the overall performance. These strategies were found to increase sensitivity to minority classes. The impact of unknown cells on semantic segmentation is demonstrated, with some evidence of the model applying learning of labelled cells to these anonymous cells. These findings indicate both classical models and new deep learning networks as promising methods in automating RBC abnormality detection.

\end{abstract}




\end{frontmatter}


\section{Introduction}
\label{sef:intro}

Red Blood Cells (RBCs) are critical in the transport of oxygen around the body. Abnormalities in the physical conformation of the cells can lead to various medical conditions due to an insufficient supply of oxygen to tissues. Typically, a Haematologist will visually examine a microscopic image to detect abnormal RBCs, a time-consuming and subjective process \citep{Hegde2018PeripheralReview,Aime2019StrategicImaging}. This paper proposes to automate this process using machine learning technologies, thus increasing throughput of disease detection, reducing diagnosis time and potentially introducing some standardisation to the practice \citep{Aime2019StrategicImaging}.

A number of studies have approached automated cell classification using machine learning models. Initially, a model based on support vector machine (SVM) was built from morphological and textural features \citep{Shirazi2018ExtremeClassification,Devi2018MalariaSmear}. Modern deep learning approaches have increasingly been used in to automate cell classification, e.g. artificial neural networks (ANNs) \citep{Tomari2014ComputerImage}, Convolutional Neural Networks (CNNs) \citep{Xu2017AAnemia,Durant2017VeryErythrocytes} and the pretrained U-Net achitectures \citep{Zhang2017ImageU-Net}. However, there are still limitations resulting low accuracy of classification, including small training datasets \citep{Devi2018MalariaSmear,Zhang2017ImageU-Net}, using imbalaned data \citep{Xu2017AAnemia,Durant2017VeryErythrocytes,Zhang2017ImageU-Net}, classifying only one disease or low number of classes \citep{Xu2017AAnemia,Zhang2017ImageU-Net,Devi2018MalariaSmear}, using a low number ($<$10) of features \citep{Tomari2014ComputerImage} or discarding overlapping cells  \citep{Durant2017VeryErythrocytes}.

Avenues for further research drawn from state of the art have been highlighted by \citet{Hegde2018PeripheralReview}. They indicated that the existing automated methods can only detect a specific disease, not yet detect a wide range of abnormalities. One of the main issues are that these methods are not robust to illumination and colour variations (see examples of various microscopic images in Table \ref{fig:celldesc}). We have echoed these points by considering a large number of abnormal cell classes and investigating a large number of features.  As the images are of varying quality, with blurring around the cell edges and variation in colour, these differences will aid the models' robustness to such variation. 

The most critical challenge is the dataset’s highly imbalanced nature, leading to rare conditions being undetected \citep{Litjens2017AAnalysis, Ker2017DeepAnalysis}. Table \ref{fig:celldesc} shows the distribution of our dataset over the 11 cell types is very uneven, ranging from 1\%-30\%. The issue with an imbalanced dataset is that the model will be very good at recognising the majority classes, as it will have been trained with a lot of information about these classes, and worse at recognising the minority classes which it has less information about. The most common techniques used in deep learning is  data augmentation in data space \citep{Anantrasirichai:Application:2018, Korranat:red:2021}. These however cannot be used with SVM or when a size of dataset is very small or the data distribution is highly skewed. In this paper, We tackle an imbalanced dataset problem with two techniques. The first one is a augmentation technique in feature space using Synthetic Minority Over-sampling Technique (SMOTE) \citep{Chawla2002SMOTE:Technique}, which to the author’s knowledge has not been applied to this particular problem. The second technique is cost-sensitive learning approaches \citep{He2009LearningData}. We also investigate the combination of these two techniques to balance the competing metrics, leading to an overall improvement of RBC classification.

\begin{table}[t!]
	\caption[Cell distribution in dataset]{Cell types, frequency and example image from our dataset.}
	\centering
	\includegraphics[width=\columnwidth]{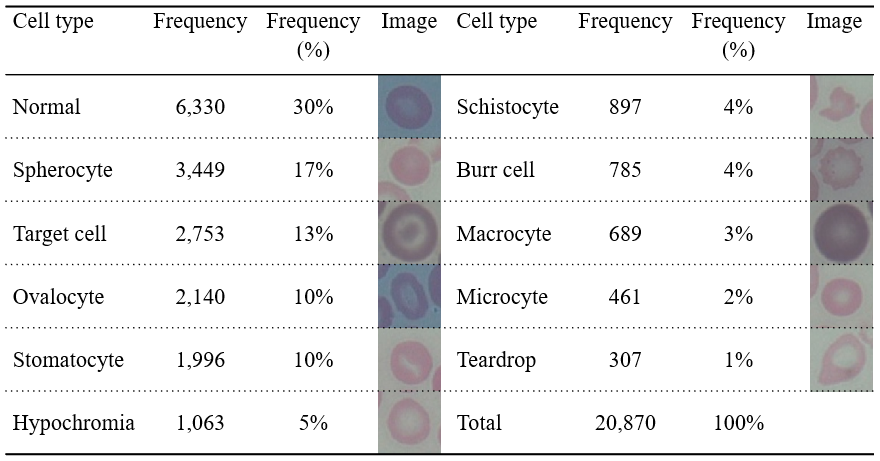}
	\label{fig:celldesc}
\end{table}

In this paper, we analysed the performance of two most potential deep learning methods: TabNet and U-Net. TabNet was introduced by Google Cloud AI \citep{Arik2019TabNet:Learning} as a deep learning architecture for tabular data. The authors assert that while tasks involving images, text and audio have harvested the benefits of deep neural networks, tasks using tabular data have not despite being the most widespread data form. U-Net \citep{RonnebergerU-Net:Segmentation} is a deep learning network first created for medical image segmentation. U-Net has been used in a wide range of medical applications including lesion and organ segmentation, cardiac image resolution and cell segmentation \citep{Litjens2017AAnalysis}. U-Net has become one of the popular architectures for semantic segmentation for other applications due to its efficiency and simplicity of implementation. 
The performance of TabNet and U-Net on RBC classification are compared with a traditional machine learning method, SVM. We also analyse features and RBC types that influence on the performances of the trained models.

The remainder of this paper is organised as follows. Dataset and data processing prepared for training the models are described in Section \ref{chap:materialsMethods}. Then data imbalance is discussed in Section \ref{sec:dataimb}. The automated methods of abnormal cell detection and RBC type classification are present in Section \ref{sec:classification}. The
performances of the methods is evaluated, compared, intensively analysed and discussed in Section \ref{chap:main}. Finally,
Section \ref{chap:conc} presents the conclusions of this work.

\section{Materials}
\label{chap:materialsMethods}

\subsection{The data}
\label{ssec:data}

Our dataset were provided by Chulalongkorn University \citep{Korranat:red:2021} with total 591 microsopic images of RBCs.  Three haematologists checked each cell and the majority vote was used as the label. If the cell was given three different labels, the cell was not included in the dataset; this was $<$1\% of all cells checked. The cells were classified into 11 classes, with 20,870 individual cells labelled. This is a higher number of classes and larger dataset than previous studies (often $<$1,000 as stated in Section \ref{sef:intro}). Broadly two thirds of cells were abnormal, split between 10 classes, and each image is of a specific disease. The 11 RBC classes (including normal cells)  with their frequency and an example image are tabulated in Table \ref{fig:celldesc}.

As the images were from different machines, it was desirable to first normalise them to reduce the differences in colour and illumination. We converted colour to grayscale and applied illumination normalisation. Note that we compared this dataset with that applied denoising and contrast enhancement. It appears that denoising removes useful information by blurring out small details in the image that may be important markers differentiating cell types. Sharpening and contrast enhancing can amplify the noise left in the images, affecting textural features. Thus the original data was used and let the machine learning models deal with these non-linear characteristics of the data, particularly the deep learning that shows the robustness of noisy data \citep{Anantrasirichai:Application:2018}.

\subsection{Ground truth generation}
\label{ssec:ex_data}
The three classification models employed in this work (SVM, TabNet and U-Net) have different labelling requirements. SVM and TabNet models take features as input, so here a region of interest (ROI) for a cell manifests in a row of data, one column of which is the label. In contrast, U-Net, a semantic segmentation approach, takes images as input, where each pixel is labelled. Thus, two different labelling approaches were taken - feature extraction and ground truth pixel labelling.

\subsubsection{Cell region identification}
\label{ssubsec:subsec01}
For both labelling approaches, the cell region first needed to be identified. At first, the label function from the Python library scikit-image \citep{VanDerWalt2014Scikit-image:Python} was used which is a connected component labelling algorithm that looks at whether neighbouring pixels have the same value. We propose a two-step approach, where the first step applies Otsu's threshold to threshold the images, separating cells (foreground) from the background, and the second step employs two iterations of open and close morphological operations.

\begin{figure}[t!]
	\centering
	\includegraphics[width=\columnwidth]{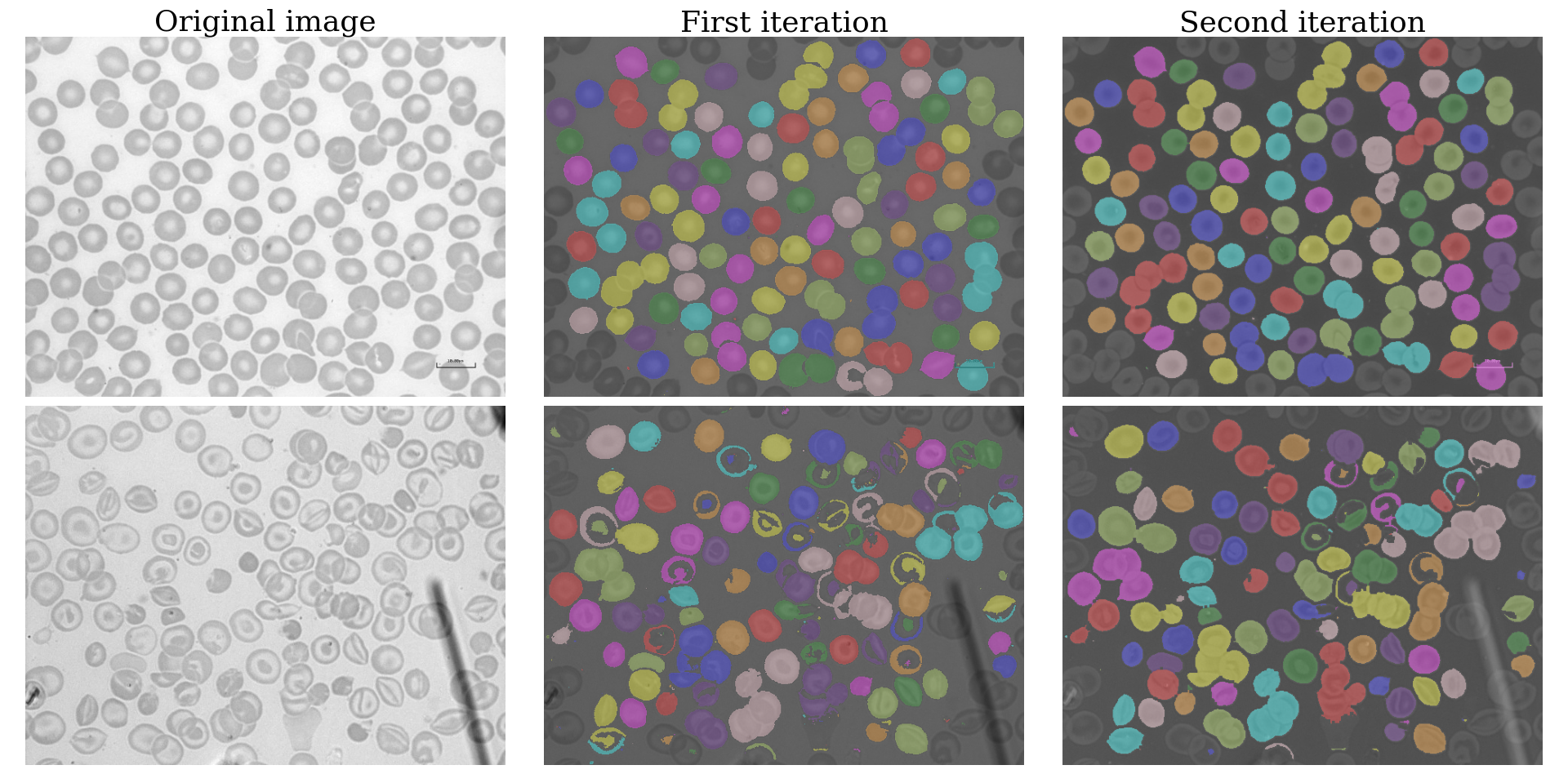}
	\caption[Examples of cell region identification]{Examples of cell region identification. First column: original images. Second column: first iteration using skimage label function with Otsu thresholding. Third column: Additional morphological operations used.}
	\label{fig:roi_id}
\end{figure}

The cell region identification algorithm performed well for cells which were well defined. However, it performed less well where cells had poor contrast to the background. This can be seen in the figure \ref{fig:roi_id}. The cells in the top image have good contrast and were picked out well in the first iteration (second column), the cells in the bottom image had poorer contrast and so not all of the cell was captured. A set of morphological operations were performed on the images and were found to enhance labelling (third column in figure \ref{fig:roi_id}). Whilst this improved cell region identification, not all cells were identified. Those with particularly low contrast required a high level of operations which then deformed cells previously well labelled so that they merged. Thus the operations used to produce the second iteration was used.

\subsubsection{Separating single and overlapped cells}
\label{subsubsec:sec2}

Overlapping cells bring more complexity to the task, obscuring key morphological and textural elements of each other, and adding information unrelated to the cell in question. Therefore overlapping cells were separated out from the database. In the literature various measures have been used to identify single and overlapping cells (area / convexity / eccentricity \citep{Wei2015AImaging,Abu-Qasmieh2018NovelCells}) but as there were more class types and very different cell shapes in this dataset, this was not sufficient for adequate identification.

We separate the overlapped cells from a single cells simply using the SVM. The training patches was constructed from 55 images - five images for each cell type with the highest number of cells of that type. Cell regions were labelled as single or overlapped, yielding 4,235 cell regions with 12.5\% labelled overlapping. The 124 morphological and textural features were extracted and used to train an SVM. The separator model was able to correctly identify 90.6\% of overlapping cells.

\subsubsection{Pixel-wise ground truth for semantic segmentation}
\label{subsec:groundtruth}

Semantic segmentation models take images as the input, with corresponding label images where pixel values represent the ground truth -- the class number is assigned to each pixel. In our labelling process, the experts provided only the centre position of the cell where they are confident. Therefore we had to create the images of ground truth from cell region identification described in Section \ref{ssubsec:subsec01}.

\begin{figure}[t!]
	\centering
	\includegraphics[width=\columnwidth]{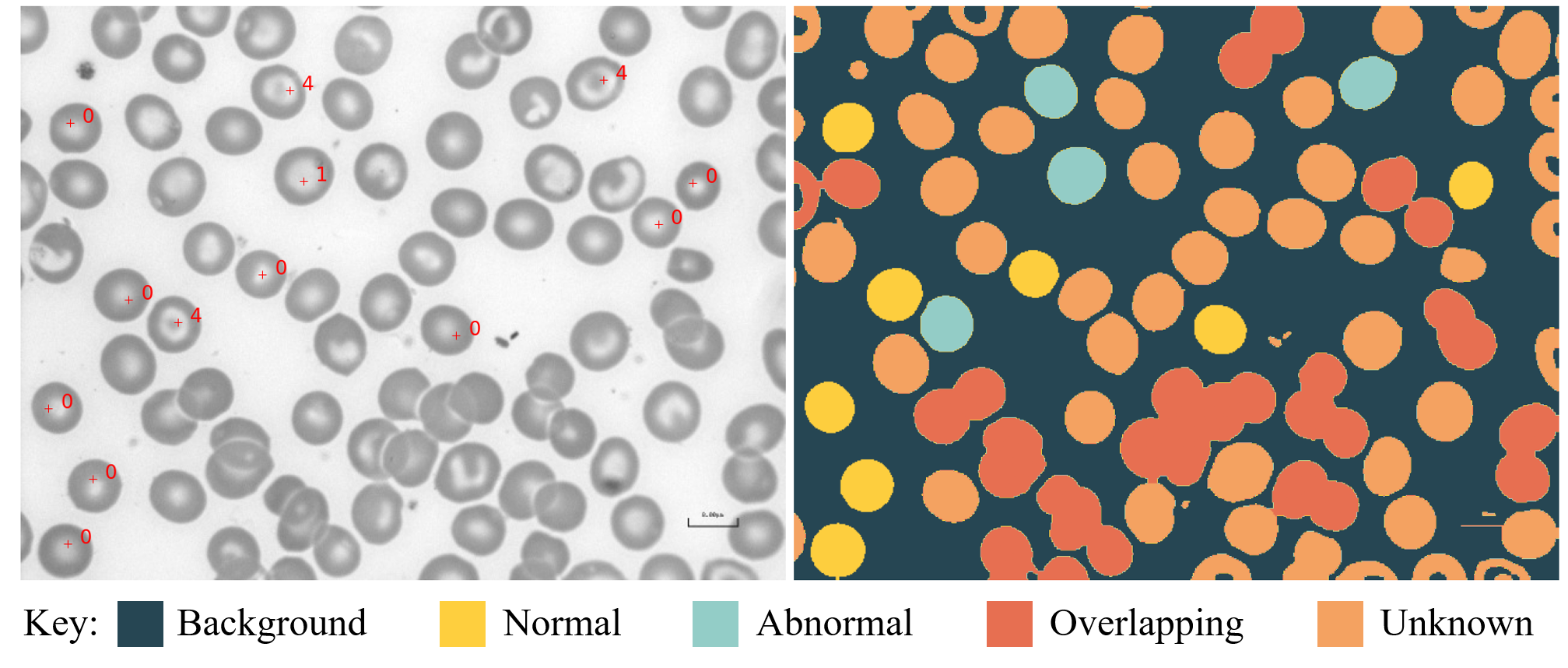}
	\caption[Ground truth labelling]{Sample used for training U-Net. Left image: Original image with ground truth labelling shown in red text. Right image: Labelled image five class version.}
	\label{fig:sslab5c}
\end{figure}

Here, two versions were created for five and nine classes. The five class version is analogous to the two class feature extraction with a normal (3\% of pixels) and abnormal (7\%) class, alongside background (66\%), overlapping (8\%) and unknown (15\%) classes. Hence this can represent an abnormal cell detection. For the nine class version the abnormal class was expanded to include four individual abnormal cell types with higher frequencies (Spherocyte (frequency in dataset 16\% in pixel manner), Stomatocyte (11\%), Target cell (11\%), Ovalocyte (10\%)).  Instead of using the whole 11 classes as in SVM and TabNet tests, the remaining six abnormal cell types were grouped together as an ‘other’ abnormal group (totalling 20\% of the dataset).
This is because the number of pixels of these individual cell types is too small. Figure \ref{fig:sslab5c} shows an example of an image with its five class labelled counterpart. The overlapping class was populated using the overlapping cell detection model, described in Section \ref{subsubsec:sec2}. The unknown class was made up of cell regions which had been identified but did not have a ground truth label. This posed a challenge to the U-Net model, as it was `seeing' images of different types of cells, which appeared in the other classes, but all had the same label. 

\section{Addressing Data Imbalance}
\label{sec:dataimb}

Imbalanced data is a common issue in biomedical data where data is skewed towards normal images with abnormal cases in the minority \citep{Gupta2019DeepReview}, affecting both classical machine learning and deep learning. Steps can be taken to mitigate the impact, either applied to the data or to the model. 

Resampling methods were found to be a common technique to address imbalanced datasets in several papers reviewing both classical machine learning \citep{Lopez2013AnCharacteristics,Iranmehr2019Cost-sensitiveMachines} and deep learning \citep{Gupta2019DeepReview,Buda2018ANetworks,Min2017DeepBioinformatics}, from simple over/undersampling to augmentation. Purely oversampling risks overfitting the model, which augmentation can protect against. The easiest form of augmentation is in data space. This consists of simple transformations via rotation, translation, mirroring, zooming and warping, and is especially popular in image classification as it is easy to implement in images \citep{DeVriesDatasetSpace}; note that in this case, changing the image size or warping the image is not appropriate as size and shape are critical RBC classification features. However, when the input is a feature set, augmenting in data space does not always provide a classifier with additional useful information as cases will be simply replicated in feature space. Instead, augmenting in feature space can be used to generate useful synthesised data examples. This has been used in many applications in classical machine learning but as far as the author is aware, not to RBC classification tasks.

Synthetic Minority Over-sampling Technique (SMOTE), one of the first augmentation techniques in feature space, was introduced by \citet{Chawla2002SMOTE:Technique} which synthesises minority class examples in feature space. The method creates synthetic examples by taking points along lines which join a minority class example and its k nearest neighbours of the same class, and was found to improve classification accuracy for minority classes. Since being introduced, SMOTE has become a popular augmentation technique and several variants have been created. 

In terms of the model itself, in classical machine learning (e.g. SVM), varying error costs were put forward as a common method to address class imbalance \citep{Batuwita2013ClassMachines}, penalising misclassification of minority classes. Similarly for deep learning models, cost functions can be applied which vary between classes. In their review of deep learning in microscopy, \citet{Xing2018DeepSurvey} stated that cost-sensitive learning had not been applied to microscopy image analysis and as far as the author is aware, this is still the case. Cost-sensitive learning in deep learning increases the impact minority classes have on both learning and network output by varying elements of the network by class-specific parameters. One can minimise the misclassification cost, rather than solely error, by adding in cost-sensitive parameters to the loss function. 

The importance of data pre-processing and data augmentation is highlighted in the review of deep learning in medical image analysis \citep{Litjens2017AAnalysis}, where the authors state that high performing algorithms were often the result of these techniques rather than the network architecture itself. We compared the impact of these strategies, namely data space augmentation, feature space augmentation and cost sensitive learning.


\subsection{Data space augmentation}
\label{subsec:subsec05}
The structure of the U-Net model requires relatively large input patches as the dimensions are reduced by a factor of 16. Here, the images were tiled into 256 $\times$ 256 squares, six for each image so that there was an element of overlap. This produced 3,546 tiles. Augmentation took the form of rotating the image three times and flipping the original image along the horizontal and vertical axis, yielding five additional images. Images to augment were chosen based on the percentage of pixels which were normal or abnormal cells. A threshold of 10\% was chosen, which meant that 36\% of image tiles were augmented. 


\subsection{Feature space augmentation}
\label{subsec:subsec04}

\begin{figure}[t!]
	\centering
	\includegraphics[width=\columnwidth]{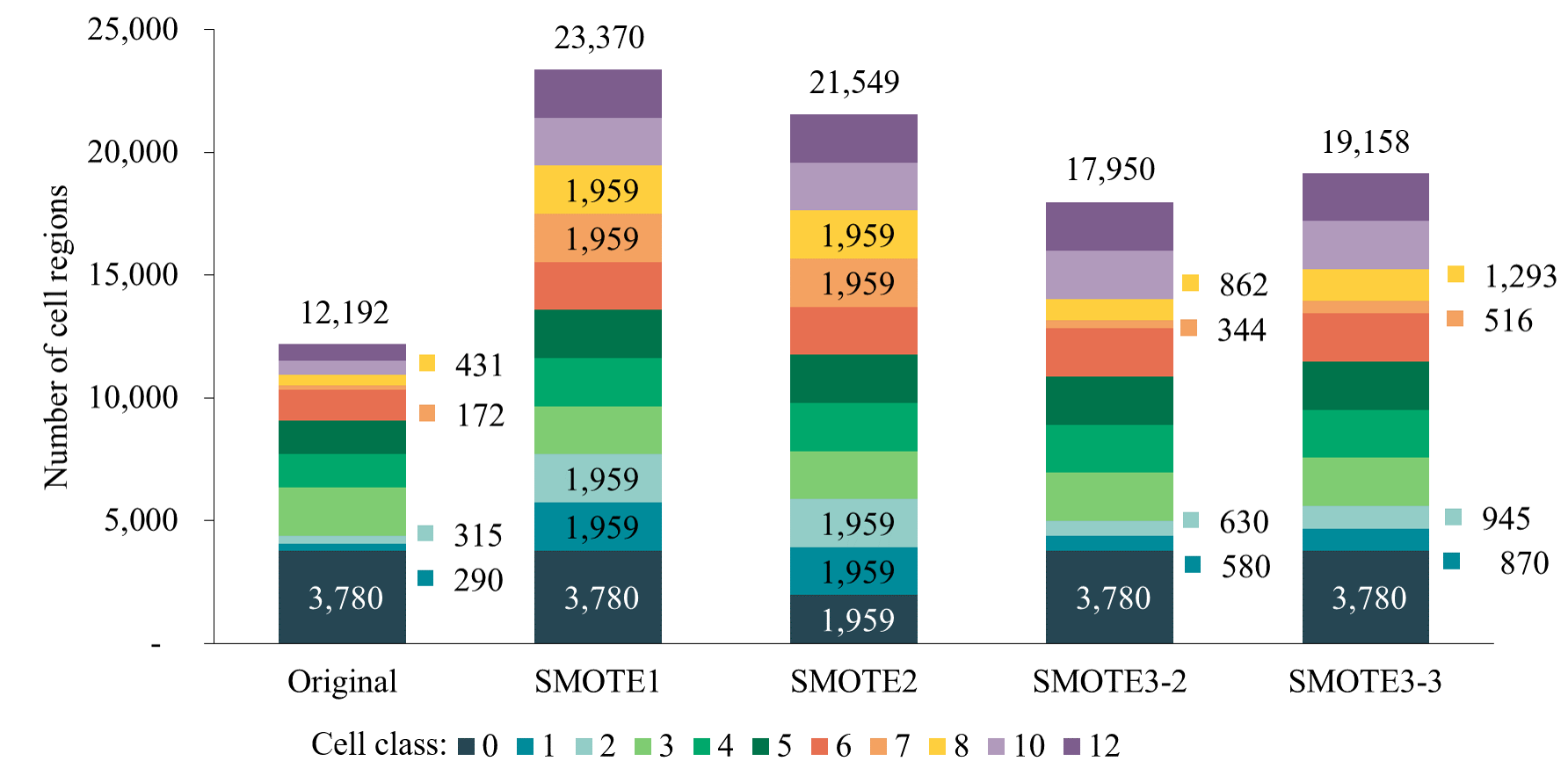}
	\caption[Effect of SMOTE on dataset size by class]{Size of classes for original training dataset and with SMOTE variations. Total size of training dataset given at top of each bar. Note SMOTE upsampling was applied to the training data only.}
	\label{fig:smote_base}
\end{figure}

We tested four variations of SMOTE, summarised in Table \ref{table:5} and visualised in Fig. \ref{fig:smote_base}, where classes 0-10 are normal cells, Microcyte, Macrocyte, Spherocyte, Target, Ovalocyte, Stomatocype, Teardrop, Burr, Hypochromia and Schistocyte, respectively. This was used on the tabular datasets. Note that in two class datasets, SMOTE was performed on an 11 class dataset and the labels then changed to two classes.

In SMOTE1 and SMOTE2 all minority classes were upsampled to the size of the largest minority class (MaxMinSize). As the data was so imbalanced, the smallest classes were being upsampled by a large amount and the training data was thus made up of largely augmented data. To check the effect of this, classes of fewer than 500 samples (classes 1, 2, 7 and 8) were instead taken to a smaller multiple of the original size in SMOTE3-2 and SMOTE3-3.
For SMOTE2 the majority class was downsampled to MaxMinSize using the NearMiss-2 algorithm \citep{Lemaitre2017Imbalanced-learn:Learning}. This keeps samples which are closest to the farthest samples of the other classes. In other words, it keeps samples along the borders to other classes, and removes samples which are easy for the model to predict. This was found to drastically reduce the ability to correctly classify normal cells and so the majority class was left unchanged in the next iterations, SMOTE3-2 and SMOTE3-3.

\begin{table}[t!]
\caption{SMOTE variations}
\scriptsize
\centering
\begin{tabular}{@{}lll@{}} 
\toprule
    \centering
    \textbf{SMOTE}  & \textbf{Majority class} & \textbf{Minority classes} \\ 
     \textbf{variation}& (Normal cells) & (Abnormal cells) \\ \midrule
    SMOTE1 & Unchanged & All taken to largest minority class (MaxMinSize) \\
    SMOTE2 & Undersampled & All taken to MaxMinSize \\
    & to MaxMinSize  & \\
    SMOTE3-2 & Unchanged & Classes $\geqslant$ 500 upsampled to MaxMinSize \\
      &  & Classes $<$ 500 upsampled to 2 $\times$ original size \\     
    SMOTE3-3 & Unchanged & Classes $\geqslant$ 500 upsampled to MaxMinSize \\
      &  & Classes $<$ 500 upsampled to 3 $\times$ original size \\
 \bottomrule
\end{tabular}

\label{table:5}
\end{table}

\subsection{Cost sensitive learning}
\label{subsec:subsec05}

The effect of cost sensitive learning was investigated for the SVM and TabNet models. In the two class case, normal cells are in the minority as abnormal cells were grouped together. Thus weighting learning using initial distribution increases the importance of learning normal classes. In fact the opposite is desired where the importance of learning abnormal cells is increased. Therefore, these weights were designed, rather than using initial distribution as is typical. In the 11 class case, abnormal cells are in the minority and so using initial distribution to weight learning was appropriate.

A cost sensitive loss function was used in the U-Net model. Cross entropy loss is the most popular loss function in image segmentation \citep{AsgariTaghanaki2020DeepReview} and with imbalanced datasets, weighted cross entropy provides cost-sensitive learning. Dice loss is also commonly used in medical image segmentation but has been found to be less stable than loss functions based on cross-entropy \citep{AsgariTaghanaki2020DeepReview}. Dice loss can be adversely affected by minority classes which are not present in each image (prevalent in this work), leading to erratic loss profiles. The two functions were compared for the 5 class model and weighted cross entropy was found to enable faster training with higher validation sensitivity to normal and abnormal cell classes. Thus weighted cross entropy was used in the U-Net models.

\section{Red blood cell classification}
\label{sec:classification}

This section presents details of automated methods implemented in this paper. Firstly the feature extraction used in SVM and TabNet is described, followed by their training settings. U-Net is a deep learning method so features are learnt from the data, so features are not required to be predefined.

\subsection{Feature extraction for SVM and TabNet}
\label{subsec:f_extraction}

As RBC abnormalities are largely identified via morphology and shading, the majority of features extracted in cell classification work are morphology and texturally based \citep{Shirazi2018ExtremeClassification}. Feature types can be broadly split into two groups – those describing the shape of the cell and those describing the texture, or shade intensity variations, of the ROI (Region of Interest). Studies have used features from either or both of these pots and the number of features extracted varies widely. It is not the case of the more features the better, and other factors such as data quality, pre-processing and model used etc. will affect a model’s success. Determining features to be used can also come from more of a haematologist’s perspective. Merino et al. \citep{Merino2018OptimizingAnalysis} set out to identify key quantitative features which characterise the morphology of different cells. This paper can be used as a sense check for the features seen in state of the art, and corresponds to what has been used.

Here, eleven morphological features were extracted for cell regions; Area, Filled area (pixels of region with holes filled), Convex area (pixels of convex hull image – a convex polygon which surrounds the cell region), Bounding box area, Solidity (Area / Convex area), Eccentricity, Extent (Area / Bounding box area), Minority and Majority axis lengths (Long and short axis of an ellipse created with equivalent normalised second moments), Axis ratio (Majority axis length / Minority axis length) and Perimeter. Five groups of texture features were extracted, generating 113 individual features: Intensity distribution, Gray Level Co-occurrence Matrix, Linear binary pattern, Grey-level run length matrix and Dual-Tree Wavelet features as used in \citep{Anantrasirichai2014Adaptive-weightedTomography} which extracted the features identified both in previous studies and from key characteristics used by haematologists.

\subsection{SVM}
\label{sec:svmmethod}

The SVM models were built using the sklearn Python library \citep{Pedregosa2011Scikit-learn:Python}. After separating the dataset into features and labels, a stratified 5-fold split was used in order to run 5-fold cross validation. The `stratified' part of the 5-fold ensures class distribution is preserved in each fold. In both the two and eleven class cases, the stratified 5-fold split was performed on the eleven class data. In the two class case this ensured the five datasets were similar, and the labels were converted to two classes after this. For each fold, the training data was normalised and the same scaling applied on the test data.  SVM models for both abnormal cell detection and RBC type classification reported in this paper used the Radial Basis Function (RBF) kernel as it outperformed the linear and polynomial kernel for all models.

\subsection{TabNet}
\label{sec:tabnetmethod}

TabNet learns the mapping based on decision trees, whilst using gradient descent-based optimisation which allows the network to learn the representations which make deep learning methods so successful. The model also quantifies how useful each feature is to the training of the model, which was found to be consistent with other feature importance ranking methods. As this model is relatively new, there are, to the author's knowledge, no studies which use it for RBC classification tasks.
\smallskip

The TabNet model was built using the pytorch-tabnet Python library \citep{Dreamquark-ai2019Tabnet}. The same steps were taken as with the SVM code with regards to the stratified 5-fold split. The TabNet model differs to the SVM model as it uses a training, validation and test dataset (rather than just train and test). The validation dataset is used to determine early stopping of training - when a certain number of epochs pass without an improvement in loss. The number of epochs to pass without improvement was set to the default of 15. Note that the TabNet model does not require data to be normalised. The hold out fold of the 5-fold split was designated as the testing dataset (as with the SVM model), the remaining 80\% of the data was then split 60:20 for the training:validation datasets.

\subsection{U-Net}
\label{sec:unetmethod}

CNNs like U-Net are made up of a number of convolutional layers and pooling layers \citep{Gupta2019DeepReview}. A convolutional layer can be thought of as a filter layer which encodes features of the image. The pooling layers downsample the convolutional layer output, achieving a reduction in the dimensions being computed whilst retaining important information. At the end of the network are fully connected layers (all neurons of layer n are connected to all neurons in layer n+1) which output class probabilities.
The detail of U-Net can be found in the supplementary material (Supp. Matt.).

The U-Net model we used was written using the Keras \citep{Chollet2015Keras}. The modification we made for RBC classification is as follows:

  \textit{Zero padding:} In the original paper the contracting side feature map was cropped because of convolution layers losing border pixels. In this work, zero padding was used so that the image dimensions were not affected.
  
  \textit{One-hot encoding: } Used for the label images, removing the problem that arises with integer labelling where classes closer together numerically are perceived to be more similar.
  
  \textit{Batch normalisation: } Not used in the original U-Net architecture, but was applied here to address the issue of internal covariate shift, caused during training as the distribution of layer inputs change as the weights of the previous layer are updated \citep{Ioffe2015BatchShift}. The model updates the weights of each layer in reverse order from the output, and each layer update assumes fixed weights in the previous layer. However, all layers are updated simultaneously. Batch normalisation enables the use of higher learning rates, thus shorter training times, as small weight changes cannot amplify sub-optimal changes in gradients. Additionally, weight initialisation had less impact on model training as the back propagation was not impacted by the weight scale.
  
  \textit{Optimiser: } Original paper used stochastic gradient descent optimiser, but here we used adaptive learning rate Adaptive Moment Estimation (Adam) optimiser.
 
A stochastic method was used with a minibatch size of 4. This relatively small size was driven by memory constraints. Five-fold cross validation was used on all models.
For five class model, due to memory constraints half of the augmented images were able to be used, a dataset of 6,756 images (up from 3,546 original images). For nine class model, memory constraints meant that no augmented images could be used. This is down to the size of the data being processed - the dataset is a four dimensional matrix the size of which is determined by the number of images, the image dimensions and the number of classes. By increasing the number of classes from five to nine the matrix almost doubled in size and so the number of images used needed to decrease.

\section{Results and discussion}
\label{chap:main}

This section details the results from the three models employed. Firstly the classification models are discussed (SVM and TabNet), and the contribution from SMOTE upsampling and cost-sensitive learning is examined. Then the semantic segmentation network U-Net is reported. The section ends with a comparison of the models.

Working with imbalanced datasets, accuracy metrics tell us very little about model performance. As normal cells are often in the majority, in an extreme case a model can achieve high accuracy scores by classifying all cells as normal. Therefore this work focuses on sensitivity, specificity, precision and the f2-score.
Rather than give performance metrics based on pixels, in this case cell based performance is of more value - how many cells were identified correctly. We extract out the cell regions of the label image to compare to the predicted image. If 
$\geqslant$ 70\% of the pixels in the predicted image had the same label as the label image, it was deemed to be a match. Cell borders were excluded from the calculation as detection, rather than exact location, was the priority.

\subsection{SVM and TabNet}

Section \ref{sec:binary} and Section \ref{sec:multi} present results of the SVM and TabNet models for two and eleven class versions. The two class version looked to classify normal and abnormal cells (abnormal cell detection), whilst the eleven class version looked to classify normal cells with the ten abnormal cell types separated out (RBC type classification).  In the binary case, abnormal cells were labelled as 1 so metrics were calculated with a true positive referring to a correctly classified abnormal cell. In the multiclass case a one against all method is used. The classes are binarised in order to calculate metrics, where the class in question is deemed positive and the remaining classes are bunched together as negative. Therefore, incorrect classifications between the negative classes are not taken into account and the number of true negatives is falsely elevated. Therefore, metrics which focus on true positives are more useful - sensitivity, precision and the f2-score.

\subsubsection{Abnormal cell detection}
\label{sec:binary}

Table \ref{table:res1} shows that the SVM and TabNet models achieved very similar performance, with f2-scores and sensitivities of approximately 92\%, and specificities of around 80\%. By increasing the number of minority class samples, SMOTE increased sensitivity to abnormal cells with corresponding decreases seen for specificity (Fig. S2 in Supp. Matt). This was a result of increases in the number of true positives with increases in false positives - cells incorrectly flagged as abnormal.
In both models SMOTE1 and SMOTE3-3 gave higher sensitivity and f2-scores whilst maintaining reasonable specificity. SMOTE2 achieved the highest gains in sensitivity however, the majority class downsampling appears to have drastically impacted the ability to correctly identify normal cells, resulting in large drops in specificity ($>$-40 percentage points (ppts)). Upsampling smaller classes ($<$500 samples) to twice their original size in SMOTE3-2, instead of three times as in SMOTE3-3, resulted in lower sensitivity gains.
Although upsampling does yield gains in sensitivity to abnormal cells, it is relatively small versus the drops in specificity - how well a model identifies normal cells.

\begin{table}[t!]
\caption{SVM and TabNet models' average performance (\%) of abnormal cell detection, CSL: Cost-sensitive learning.}
\footnotesize
\centering
\begin{tabular}{@{}lccccc@{}} \toprule
    \centering
  &	Sensitivity &	Specificity &	f2-score \\ \midrule
SVM 	& 93.6	& 79.4	& 93.1 \\
{SMOTE1+CSL(1:2)} & 98.2 &	55.7 &	94.8 \\
{SMOTE1+CSL(2:3)}  &  97.2 &	65.3 &	94.8 \\
{SMOTE3-3+CSL(1:2)}  &  98.0 &	59.1 &	94.9 \\
{SMOTE3-3+CSL(2:3)}  &  96.9 &	68.3 &	94.8 \\ \midrule
TabNet 	& 92.4 &	81.2 &	92.3 \\ 
{SMOTE1 + CSL(1:2)} & 96.6 &	63.7 &	94.1 \\ 
{SMOTE3-3 + CSL(1:2)} & 96.5 &	65.6 &	94.2\\ 
\bottomrule
\end{tabular}
\label{table:res1}
\end{table}

For case-sensitive learning, normal cells were in the minority as abnormal cells were grouped together (31\% normal, 69\% abnormal). Thus weights were designed rather than using initial distribution as is typical. Weighting schemes of normal: abnormal of 1:2 and 2:3 were tested. 
Like the upsampling schemes, cost-sensitive learning improved the model sensitivity to abnormal cells, but with a larger cost to specificity (Fig. S5 in Supp. Matt). Reducing the weighting to 2:3 helped preserve specificity much more for the SVM model than the TabNet.

The combinations of SMOTE and case-sensitive learning for SVM and TabNet classifiers increased the f2-score and sensitivity to above where a single strategy achieved.
The highest sensitivity is from the SVM model trained with the combinations with 1:2 weighting - where the model correctly identifies 98\% of abnormal cells. However, the weighting is too small for the normal class - specificity is only 56\% and 59\% (SMOTE1 and SMOTE3-3 respectively) which is too low, rendering the models not fit for purpose. The best overall performance comes from the SVM model with SMOTE3-3+CSL(2:3) weighting, achieving a sensitivity of 96.9\% with specificity of 68.3\% and f2-score of 94.8\%.


Identification of abnormal cells is key to disease detection. However, low specificity or precision resulting from higher rates of false positives - high numbers of normal cells being misclassified as abnormal - end up confusing an image's results. The choice of model needs to balance these two competing ideals and here clinical knowledge would be needed. 
Take for example the binary SVM model, a question is raised whether it is better to correctly identify 93.6\% of abnormal and 79.4\% of normal cells (original model), or 96.9\% of abnormal and 68.3\% of normal cells?

\subsubsection{RBC type classification}
\label{sec:multi}

Table \ref{table:res2} shows performance by cell type for the two multiclass models, where the average metrics show that the SVM  slightly outperforms TabNet. The multiclass metrics are lower than the binary as it is a harder task for models to classify 11 classes rather than 2; f2-scores and sensitivities were down to around 70\%. The impact of class imbalance is made clear, with the models achieving higher metrics for the more frequent classes. Broadly, the SVM model metrics were above the TabNet model, with the exception of Macrocytes and Microcytes. The SVM classifier's performance was notably higher for Teardrop cells, Hypochromia and Stomatocytes. This could suggest different strengths for the models, with TabNet being better able to differentiate between size (the defining feature of Macrocytes and Microcytes), and the SVM being more attuned to texture differences (Hypochromia and Stomatocytes). 

\begin{table}[t!]
\caption{Multiclass SVM and TabNet model performance by class.}
\footnotesize
\centering
\begin{tabular}{@{}lcccccc@{}} \toprule
    \centering
Class & \multicolumn{2}{c}{Frequency (\%)} & \multicolumn{2}{c}{Sensitivity} & \multicolumn{2}{c}{f2-score} \\ 
 &	SVM	& TabNet	& SVM	& TabNet	& SVM	& TabNet \\\midrule
Normal	& 31	& 30	& 88.3	& 83	& 86.7	& 82.0 \\
Macrocyte	& 2	& 3	& 63.1	& 77.7	& 64.2	& 77.9\\
Microcyte	& 3	& 3	& 43.0	& 50.3	& 47.0	& 52.8\\
Spherocyte	& 16	& 16	& 93.0	& 88.4	& 92.4	& 88.4 \\
Target cell	& 11	& 11	& 84.7	& 81.8	& 84.3	& 81.9 \\
Stomatocyte	& 11	& 11	& 87.4	& 77.5	& 84.9	& 76.6 \\
Ovalocyte	& 10	& 10	& 83.9	& 83.3	& 84.6	& 82.7 \\
Teardrop	& 1	& 1	& 44.7	& 30.6	& 48.8	& 33.0 \\
Burr cell	& 4	& 4	& 61.6	& 57.3	& 65.6	& 58.7 \\
Schistocyte	& 5	& 5	& 72.9	& 69.3	& 74.4	& 69.8 \\
Hypochromia	& 6	& 6	& 60.4	& 52.3	& 63.3	& 54.5 \\ \midrule
Average & - & - & 71.2 &  68.3 & 72.9 &  68.9\\
\bottomrule

\end{tabular}
\label{table:res2}
\end{table}

The SMOTE strategies increased the sensitivity of the models to minority classes, in particular for the smallest classes; Macrocytes, Microcytes, Teardrop, Burr cells and Hypochromia (see Fig. S3 and S4 in Supp. Matt).
The downsampling performed in SMOTE2 led to large drops in performance. Sensitivity to normal cells fell drastically. Interestingly, precision of the smallest classes also saw large drops, signifying an increase of false positives - perhaps as the model struggled to correctly identify normal cells.
Comparing the other three strategies, SMOTE1 yielded the largest increases in sensitivity and the f2-score, perhaps as this variety upsampled the smallest classes (Macrocytes, Microcytes, Teardrop and Burr cells) to the largest minority class size, as opposed to multiples of their original size as in SMOTE3-2 and SMOTE3-3.
The SVM model saw greatest performance improvement for Macrocytes and Microcytes, whereas the TabNet model's performance increased most for Teardrop, Burr cells and Hypochromia.

For cost-sensitive learning, the pattern of change in metrics was similar to that seen in the SMOTE variations (see Fig. S6 in Supp. Matt.). Another similarity was that the SVM model saw greater performance improvements for Macrocytes and Microcytes, and the TabNet model saw larger increases for Teardrop and Burr cells. The sensitivity gains made in the SVM model came at a higher cost to precision than those seen for the TabNet model. There were substantial decreases in sensitivity to normal cells for both the SVM and TabNet models, larger than seen with the SMOTE variations (bar SMOTE2).

The impact of the combinations on performance of SVM and TabNet can be seen in Table \ref{fig:cs_mc}. For SVM, the combinations gave very similar results, with the SMOTE1 combination just above for metrics in the majority of classes. There is a decrease for normal cells, alongside substantial gains for the smallest minority classes; Macrocytes, Microcytes, Teardrop, Burr cells and Hypochromia. There is little impact on the remainder of the minority classes, and positive to note that they did not mirror normal cells and decrease more than a couple of percentage points.

For multiclass TabNet, the SMOTE1+CSL model had a more positive impact on the f2-score than the SMOTE3-2+CSL combination. Both increases were bigger, and decreases smaller. This effect plays out in larger increases to sensitivity and smaller decreases in precision.
The SMOTE1 and cost-sensitive learning model saw substantial jumps for Macrocytes, Microcytes, Teardrop, Burr cells and Hypochromia, often more than either strategy achieved by itself. Interestingly, it appears that the combination also tempered the decreases seen in f2-score with cost-sensitive learning in particular. As SMOTE1 increased all minority classes to the same level, the cost-sensitive learning algorithm does not generate as large weights for the minority classes. With cost-sensitive learning alone, some of the classes would have been given very large weights due to the large imbalance between classes, resulting in considerably lower importance given to the larger minority classes. The combination thus appears to be beneficial as it combines the strength of the increases in performance for smaller classes, whilst dampening the decreases in performance seen in the larger classes.

\begin{table}[t!]
	\caption[CS mc combo]{Effect of combination strategies on multiclass SVM and TabNet performances, CSL: Cost-sensitive learning.}
	\centering
	{\footnotesize SVM }
	\includegraphics[width=\columnwidth]{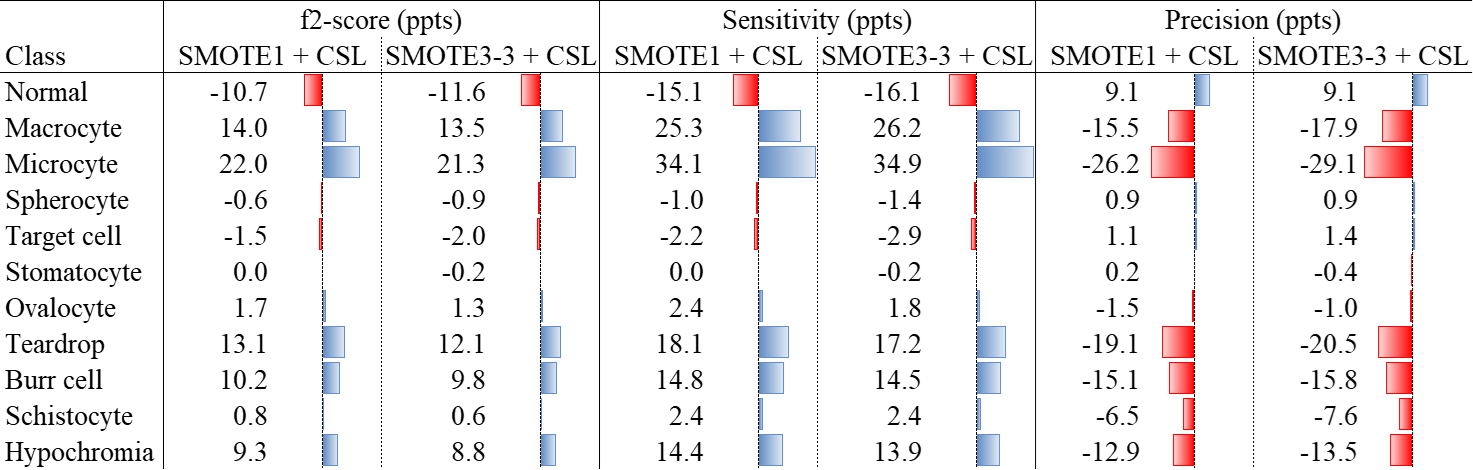}
	{\footnotesize TabNet}
	\includegraphics[width=\columnwidth]{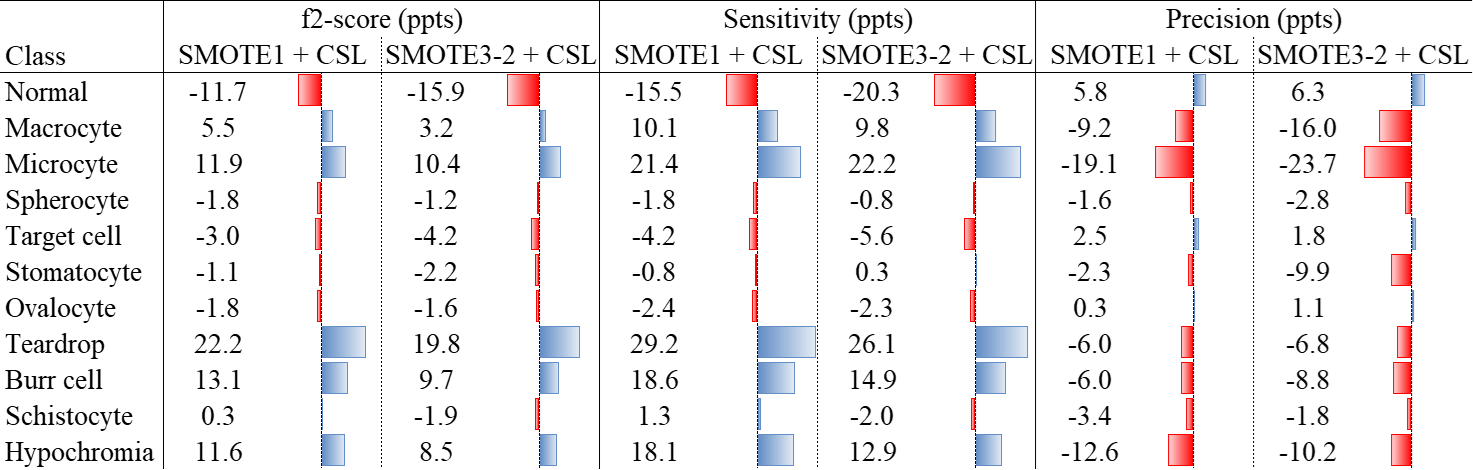}
	\label{fig:cs_mc}
\end{table}

{Common misclassifications} occurs due to small class sizes and similarities between cell types:
     i) \textit{Teardrop cells misclassified as Ovalocytes and Schistocytes}. These were the most common misclassifications for both models. Teardrop cells were the smallest class, making up just 1\% of the dataset. Both Teardrop cells and Ovalocytes have an elongated shape, and the irregular shapes of Schistocytes may explain the confusion;
    ii) \textit{Normal cells misclassified as Stomatocytes}, which have a similar size and shape;
    iii) \textit{Microcytes misclassified as Spherocytes}, both cell types have a smaller size than normal cells;
    iv) \textit{Burr cells misclassified as normal}. This was an unexpected misclassification as Burr cells have a very distinctive shape. However they do have a similar size and texture within the cell to normal cells. These were also one of the smallest classes, making up 4\% of the dataset;
    v) \textit{Hypochromia misclassified as Target cells and Stomatocytes}. These cells have the same size and shape.  with just textural differences;
    vi) \textit{Schistocytes misclassified as a range of cells}, perhaps as their fragmented nature means the models learnt this class as having a range of sizes and shapes.

\subsubsection{Feature Importance}
\label{subsec:tab_importance}

\begin{figure}[t!]
	\centering
	\includegraphics[width=\columnwidth]{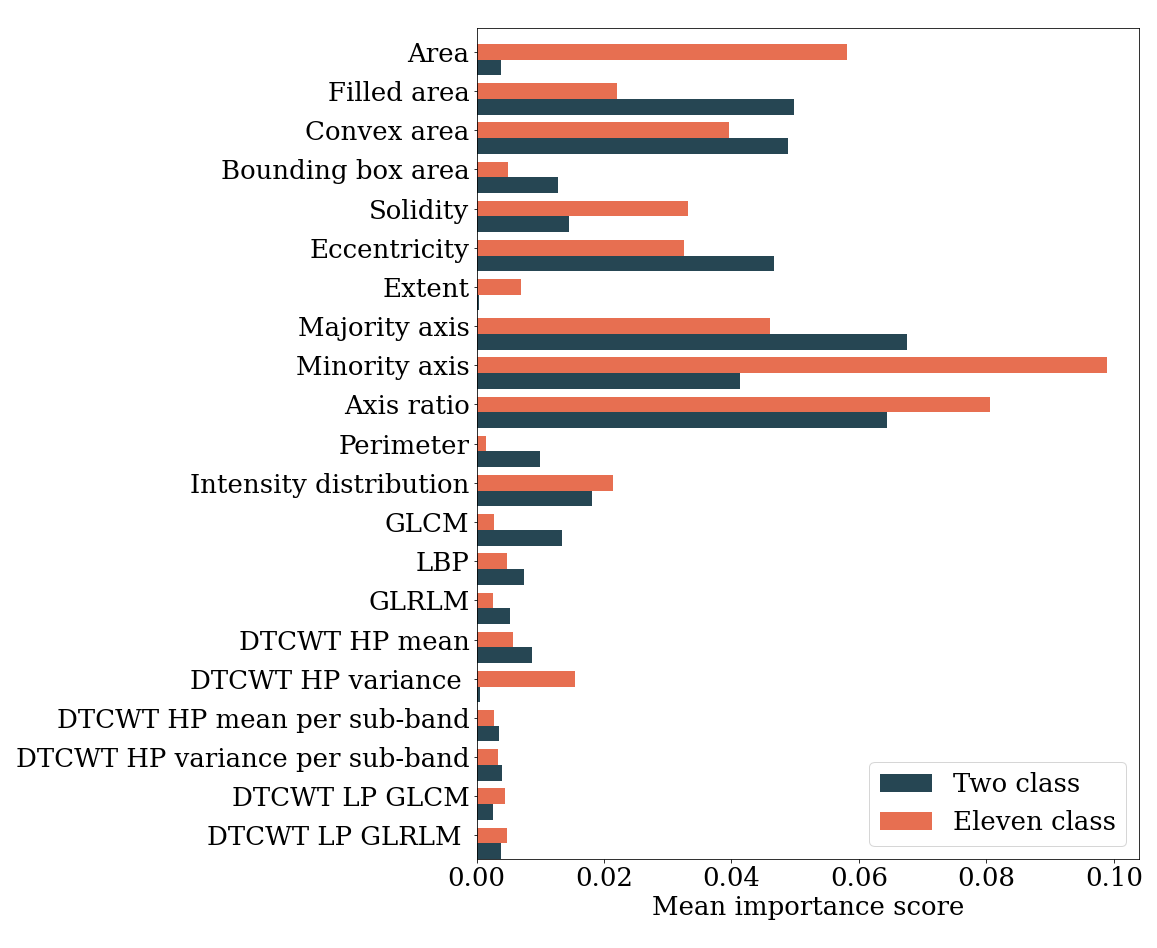}
	\caption[Mean feature importance scores for the two and eleven class TabNet model]{Mean feature importance scores for the two and eleven class TabNet model.}
	\label{fig:tabnet_f_compare}
\end{figure}

In this section, we analysed the feature importance used in TabNet algorithm.
Figure \ref{fig:tabnet_f_compare} shows the mean importance scores for the two models. For both the binary and multiclass casees, highest importance was attributed to the morphological features (the size and shape of cells) over the textural features (the pattern of grayscale intensity). Features which would be more useful in differentiating between the abnormal cell types appear to have been given higher importance in the 11 class case compared to when the model `saw' all abnormal classes as one class. These included solidity, a measure of irregular shape, and the ellipse axes, a measure of how elliptical a shape is.

Although importance ratings changed between the 5-fold cross validation runs, it was observed that importance was conserved within meaningful groups such as area or shape features. For example if the filled area was not rated highly, then the convex area was. Averaging within the groups both aided comparison (rather than analysing 124 features), but also captured this moving of importance within groups.

\subsection{Semantic segmentation with U-Net}
\label{sec:semantic_seg}

 Two tests were set, five and nine classes as described in Section \ref{subsec:groundtruth}. The confusion matrices were based on the average of the 5-fold cross validation confusion matrices. The  percentages of the true class were computed, to enable easy comparison across classes, so that rows sum to 100. The metrics are cell-based, rather than pixel-based.

We tested three fixed learning rates (0.01, 0.001 and 0.0001) using the stochastic gradient descent (SGD) optimisation algorithm as used in original U-Net. We however found that Adam (initial learning rate of 0.001) outperformed the fixed learning rate algorithms in both the five and nine class case, with a lower and more steady validation loss profile. 
In the five class case, the validation loss started to climb after around epoch 15, indicating overfitting past this point. In the nine class case epoch 20 was indicated as a good stopping point. As would be expected, the models with a lower loss profile resulted in higher validation sensitivities to pixel types, both when examining for normal / abnormal in the five class case, and the six cell types in the nine class case. We also tested several setting for dropout, but we found insignificantly different in performance. 

\subsubsection{Five class model as for abnormal cell detection}
\label{subsec:5p}

The confusion matrix is shown in Table \ref{fig:unet2c_cm}, calculating in the number of cells detected. 7\% of abnormal cells were misclassified as normal, and a further 7\% were misclassified as overlapping. Qualitatively it was seen that the algorithm which labelled overlapping cells could misclassify if cells were particularly large or elongated, which could perhaps explain why only 3\% of normal cells were misclassified as overlapping. Interestingly, 18\% of cells labelled as unknown were predicted to be normal cells, and 40\% were predicted to be abnormal cells. These predictions give a ratio of 31\% : 69\% which is very close to the ground truth ratio (30\% : 70\%). The average accuracy, sensitivity and precision are shown in Table \ref{table:bench} and f2-scores are shown in Fig \ref{fig:mc_compare_cells}.

\begin{table}[t!]
	\caption[U-Net 5 class: Confusion matrix]{Confusion matrix for five class U-Net model. Background is marked as class 5 and not included in this confusion matrix.}
	\centering
	\includegraphics[width=0.75\columnwidth]{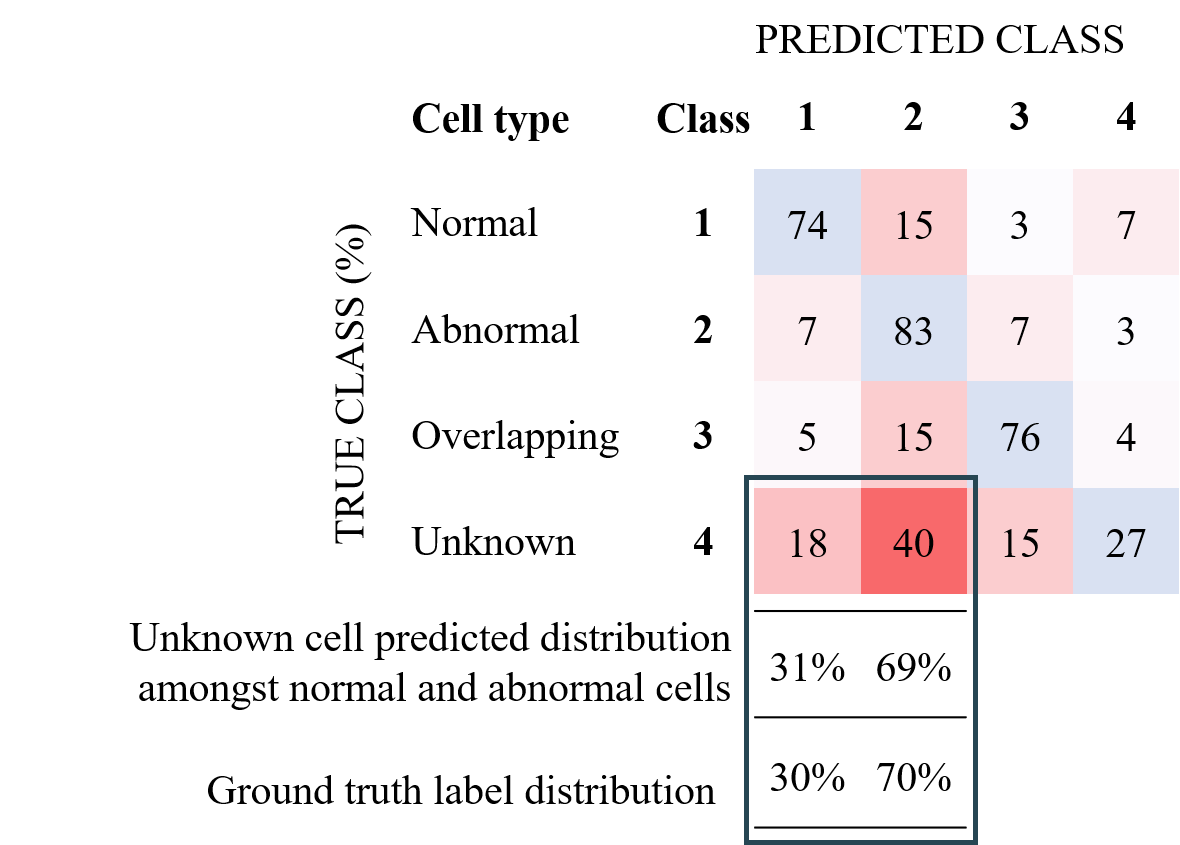}
	\label{fig:unet2c_cm}
\end{table}

\subsubsection{Nine class model as for RBC type classification}
\label{subsubsec:9c_pred}

\begin{figure}[t!]
	\centering
	\includegraphics[width=\columnwidth]{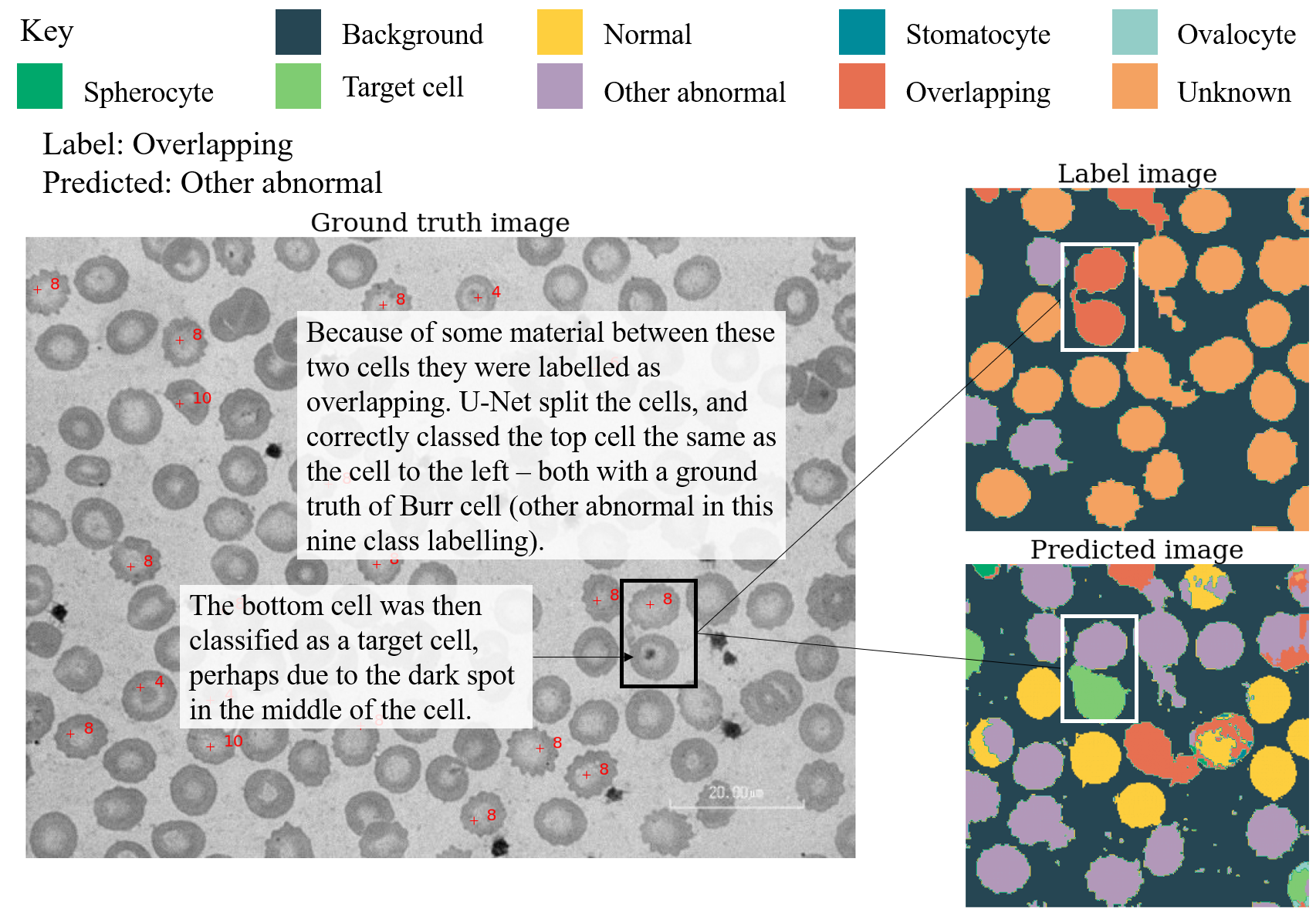}
	\includegraphics[width=\columnwidth]{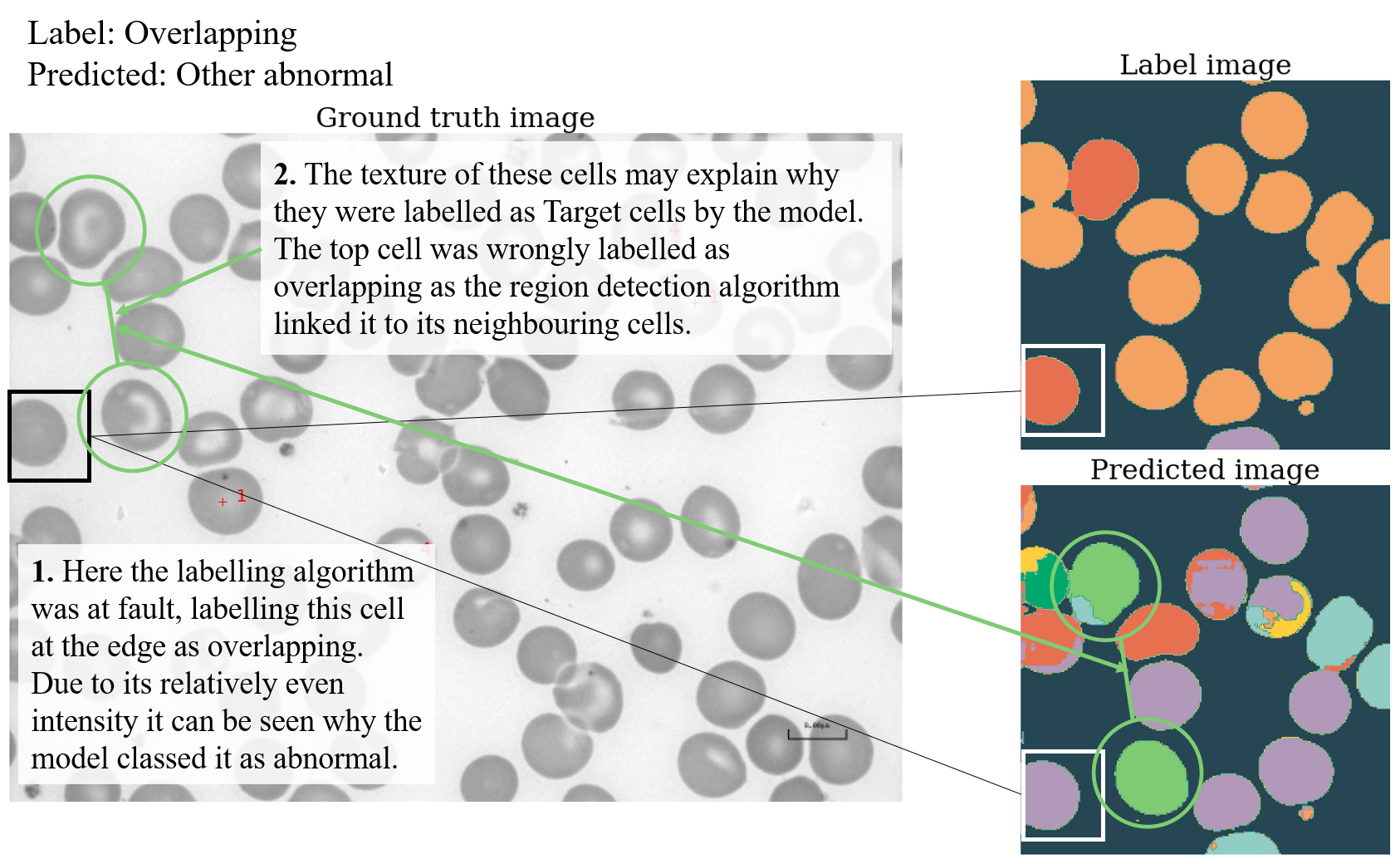}
	\caption[Examples of misclassification 1]{Cells labelled overlapping classified as other abnormal.}
	\label{fig:unet9miss1}
\end{figure}

The confusion matrix in Table \ref{fig:unet9c_cm} shows how the different cell types were misclassified.
Amongst the known abnormal classes, the bulk of misclassifications were as normal (particularly Ovalocytes, Stomatocytes and 'other' abnormal), overlapping (Target cells and 'other' abnormal cells) or 'other' (Ovalocytes). There was less confusion between the four individual abnormal cell types.
Almost a tenth (9\%) of normal cells were misclassified as 'other' abnormal, with 12\% misclassified in the four individual abnormal cell types. Interestingly the unknown class was predicted as range of cell types. Taking the distribution of the unknown cells predicted to be in the known classes (classes 1-6, see box at bottom of figure \ref{fig:unet9c_cm}), 28\% were predicted to be normal - similar to the 30\% figure based on all ground truth labels. The percentages in the four individual abnormal cell types were lower than the ground truth. The highest rate of misclassification was for overlapping cells classed as other abnormal, and Ovalocytes classed as normal (both 10\%). Fig. \ref{fig:unet9miss1} show examples of this with the original image, label image and predicted image. Despite these being picked as examples of the common misclassifications, one can still see examples within these images of the model classifying cells correctly (matching the ground truth label), as well as classifying unknown cells correctly. 

The model is most sensitive to Stomatocytes, Spherocytes and Target cells, achieving sensitivities of 82.3-83.8\%. There is lower sensitivity to Ovalocytes (72.4\%) - 10\% were misclassified as normal, the highest of the abnormal cells, and 7\% were misclassified as 'other' abnormal. The model correctly classified 72.5\% of normal cells. The performance metrics calculated with and without the unknown and overlapping cells are shown in Supp. Matt. (Table S2).

\begin{table}[t!]
	\caption[U-Net 9 class: Confusion matrix]{Confusion matrix for nine class U-Net model. Background is marked as class 9 and not included in this confusion matrix.}
	\centering
	\includegraphics[width=\columnwidth]{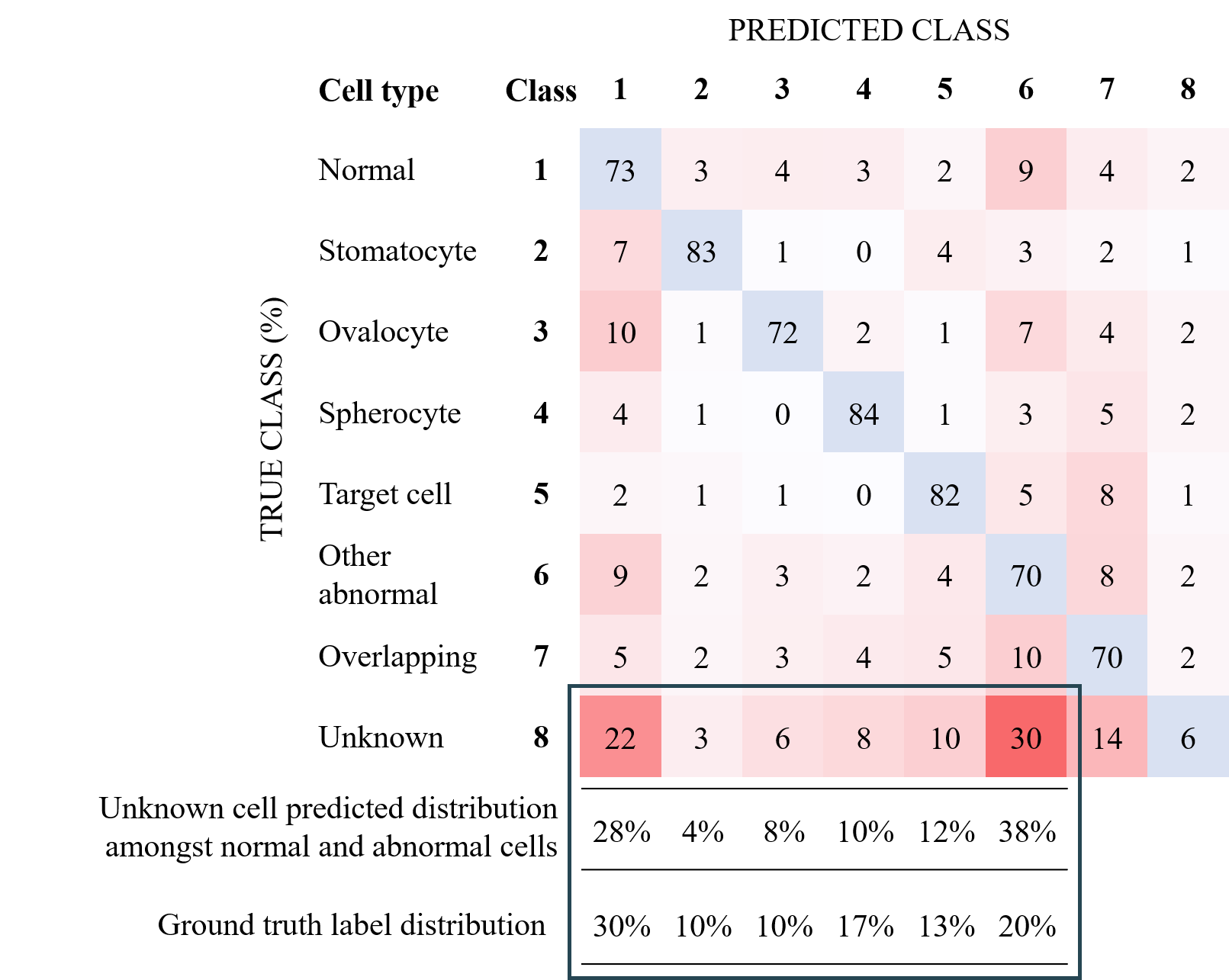}
	\label{fig:unet9c_cm}
\end{table}

The unknown cells in the images posed a challenge to the models. They made up 15\% of pixels in the dataset, more than all the known cells combined (10\%). A sample of predictions made by the network was checked to the Haematologists, images with a high number of unknown cells were chosen.

\begin{figure}[t!]
	\centering
	\includegraphics[width=\columnwidth]{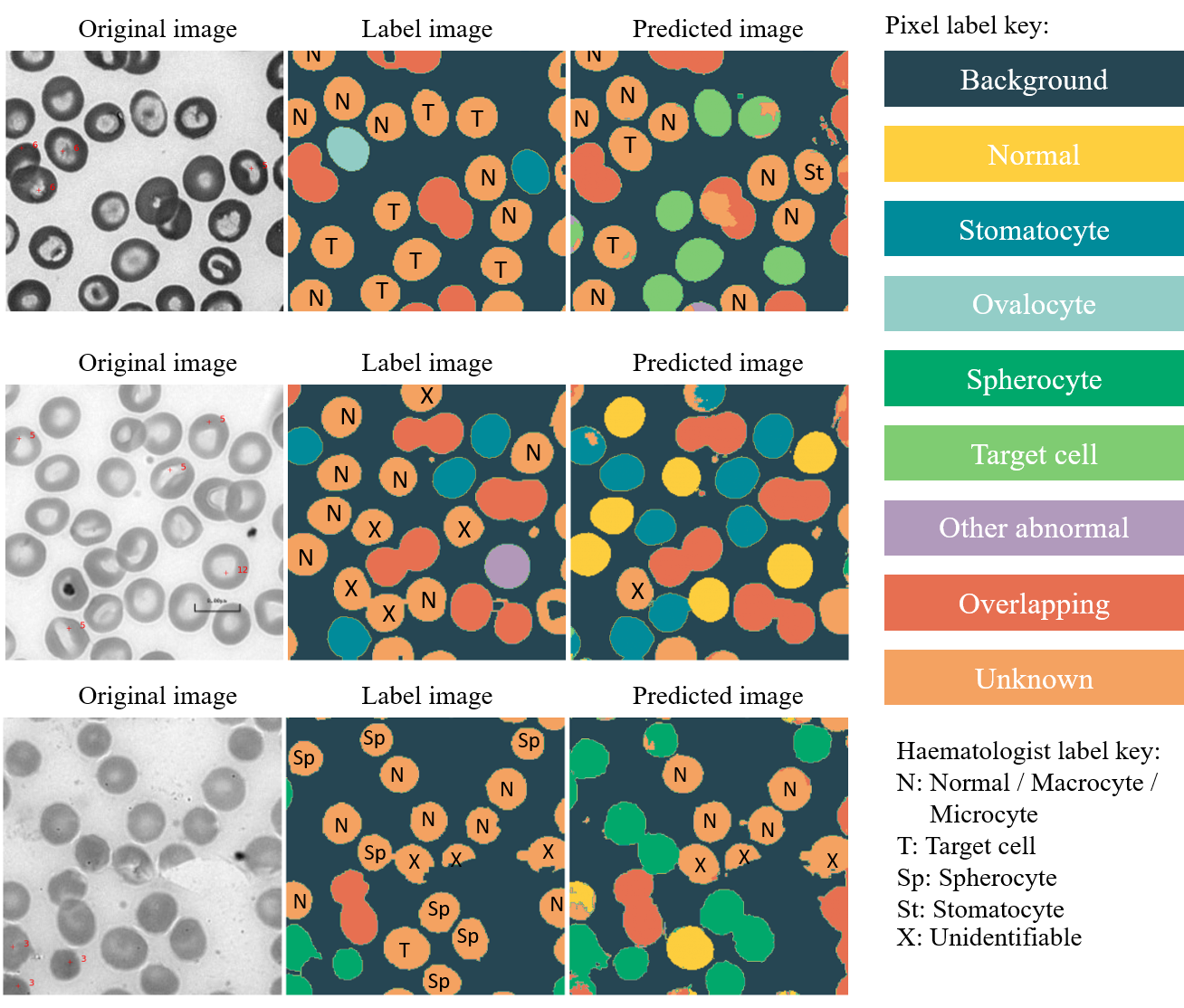}
	\caption[U-Net 9 class: Predictions with Haematologist labelling of unknown cells]{Original image, label image and predicted image for nine class U-Net model with Haematologist labelling of unknown cells.}
	\label{fig:unet9cthailabel}
\end{figure}

The original image, label image and predicted image are shown in Fig. \ref{fig:unet9cthailabel}, with Haematologist labelling overlaying the images in text. Note that due to changes in scale, N could represent normal, Microcyte or Macrocyte cells.
In the top row the model correctly identified six Target cells which had been labelled as unknown. In the middle row six normal cells were correctly classified (with the caveat that these could be Microcytes or Macrocytes). However, one cell was misclassified as Stomatocyte (upper left quadrant) and an `other' abnormal was misclassified as normal (lower right quadrant). In the bottom row seven cells were correctly classified as Spherocytes and there is a start of a correct classification of a normal cell (left border). However one Target cell was misclassified as normal and a normal cell was misclassifed as a Spherocyte.

Given the anecdotal evidence of the re-marked examples of unknown cells and the split of unknown cells between normal and abnormal classes, it is proposed that if the unknown and overlapping cells were labelled and separated, performance metrics for the model would improve. Additionally, improvements in performance of the labelling algorithm with respect to overlapping cells may well improve performance further.

\subsection{Comparison}
\label{sec:comparison}

This section compares the three models built in this paper, first for the two class case, then for multiple classes. The contribution of SMOTE upsampling and cost-sensitive learning is assessed, as well as how the models compare to state of the art.

\subsubsection{Abnormal cell detection}
\label{subsec:comparison2}

\begin{figure}[t!]
	\centering
	\includegraphics[width=0.95\columnwidth]{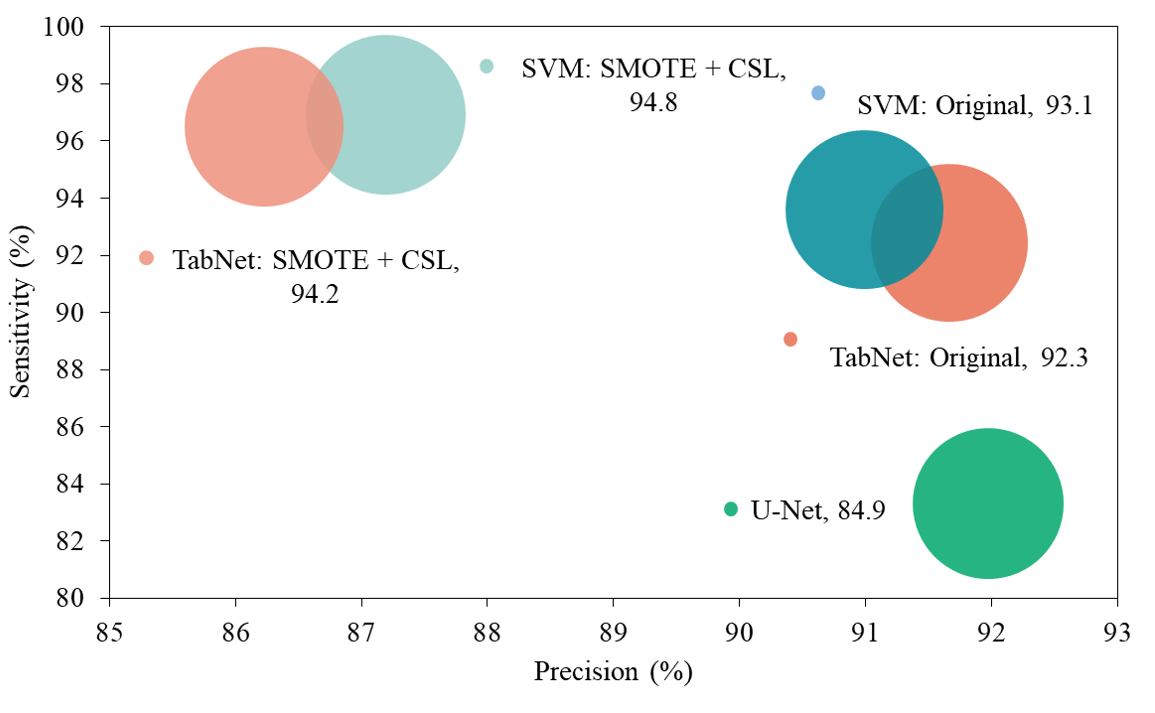}
	\caption[Comparison of two class model metrics]{Comparison of two class SVM, TabNet and U-Net. Metrics are for abnormal cells. f2-score displayed in bubble. Upsampled versions: SVM - SMOTE3-3 + (2:3); TabNet - SMOTE3-3 + (1:2). U-Net model using Dropout1, stopped at epoch 12.}
	\label{fig:2c_compare}
\end{figure}

Versions of the three models are displayed in figure \ref{fig:2c_compare}. The SVM and TabNet models are represented both using the original data, and upsampled data with cost-sensitive learning (CSL). The metrics for the U-Net model were calculated in the same way as those for the SVM and TabNet models, with abnormal designated positive, only considering known, single cells. The figure plots sensitivity against precision, with labels showing the model and f2-score. 

The highest f2-score was achieved by the SVM model with SMOTE and cost-sensitive learning.  The SVM and TabNet models utilising SMOTE and cost-sensitive learning achieved the highest sensitivities, able to correctly identify 96.9\% and 96.5\% of abnormal cells. The increases in sensitivity for the SVM and TabNet models provided by the upsampling and cost-sensitive learning came at a cost to precision - a measure of how impaired the results are by false positives (normal cells misclassified as abnormal). Whilst the U-Net model did not reach the same levels in sensitivity or f2-score, it was the most precise model - 92.0\% of the cells it identified as abnormal were abnormal. However, of the abnormal cells it could only identify 83.3\% correctly.

\subsubsection{RBC type classification}
\label{subsec:comparison11}

\begin{figure}[t!]
	\centering
	\includegraphics[width=\columnwidth]{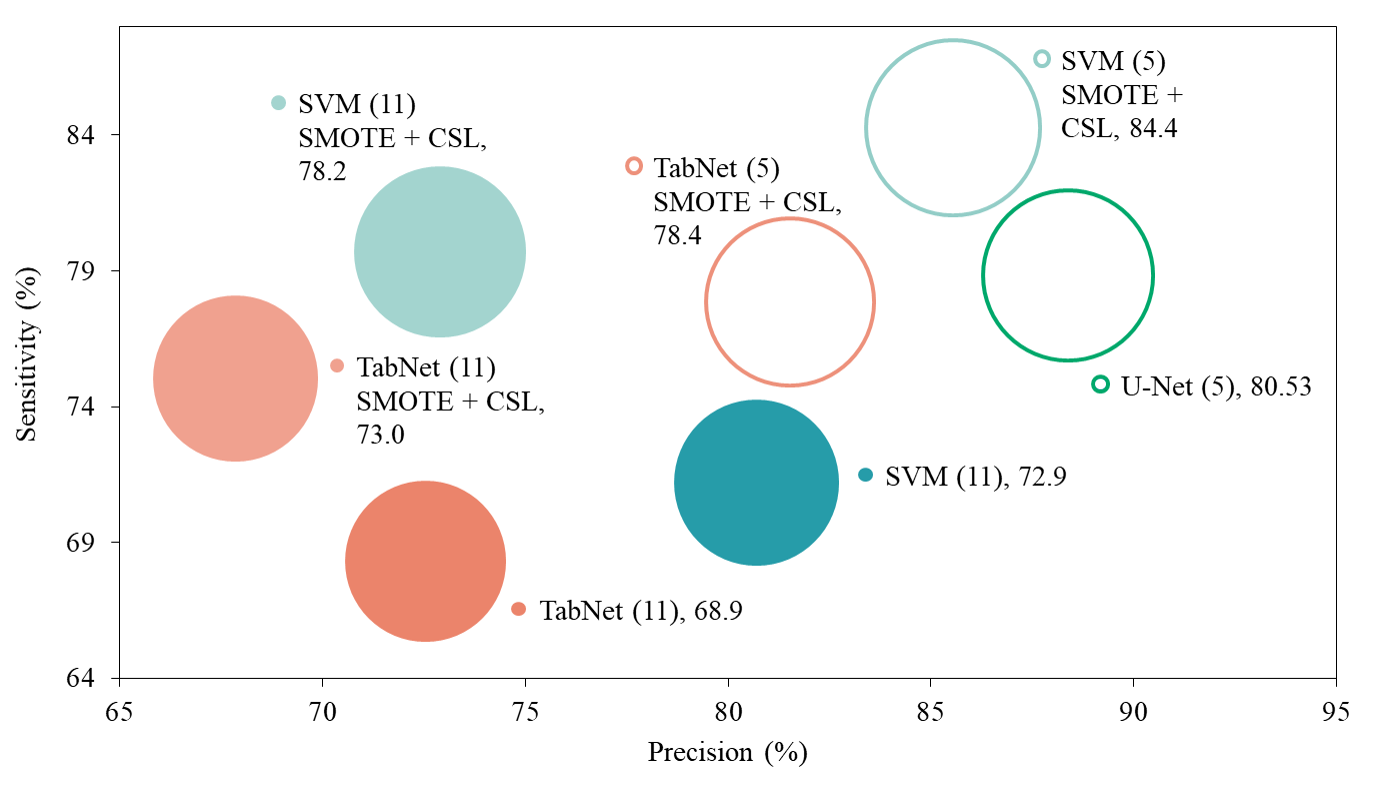}
	\caption[Comparison of multiclass model metrics]{Comparison of multiclass SVM, TabNet and U-Net. Filled circles 11 classes, open circles 5 classes. Number of classes also in brackets in each label.	f2-score displayed in bubble. Upsampled versions: SVM - SMOTE1 + CSL; TabNet - SMOTE1 + CSL. U-Net model stopped at epoch 20.}
	\label{fig:mc_compare}
\end{figure}

Figure \ref{fig:mc_compare} shows the average metrics for the multiclass models. In order to compare models differing in the number of classes they classify, the classification model metrics were also calculated as the U-Net ones were - metric averages of the five known classes (Normal, Stomatocyte, Ovalocyte, Spherocyte, Target cell). These five class metrics are represented by the open circles, whilst the 11 class metrics are represented by the filled circles. The models are plotted on precision and sensitivity, with the f2-score labelled by each circle. 

The SVM model appears to be more precise than the TabNet model in the multiclass case, in all variations seen on the plot. The impact of upsampling and cost-sensitive learning is clear, with a greater impact on the SVM model. Within the five class models, the SVM model achieves the highest f2-score and sensitivity, whilst the U-Net model is the most precise. Comparing the eleven and five class cases of the classification models also highlights that when the smaller classes are removed, both precision and sensitivity increase. These five classes make up almost 80\% of the dataset (see section \ref{subsec:f_extraction}). This reiterates the impact of imbalanced datasets on machine learning. Whilst the upsampling and cost-sensitive techniques did make a substantial impact there remains an adverse effect from these smaller classes.

\begin{figure}[t!]
	\centering
	\includegraphics[width=\columnwidth]{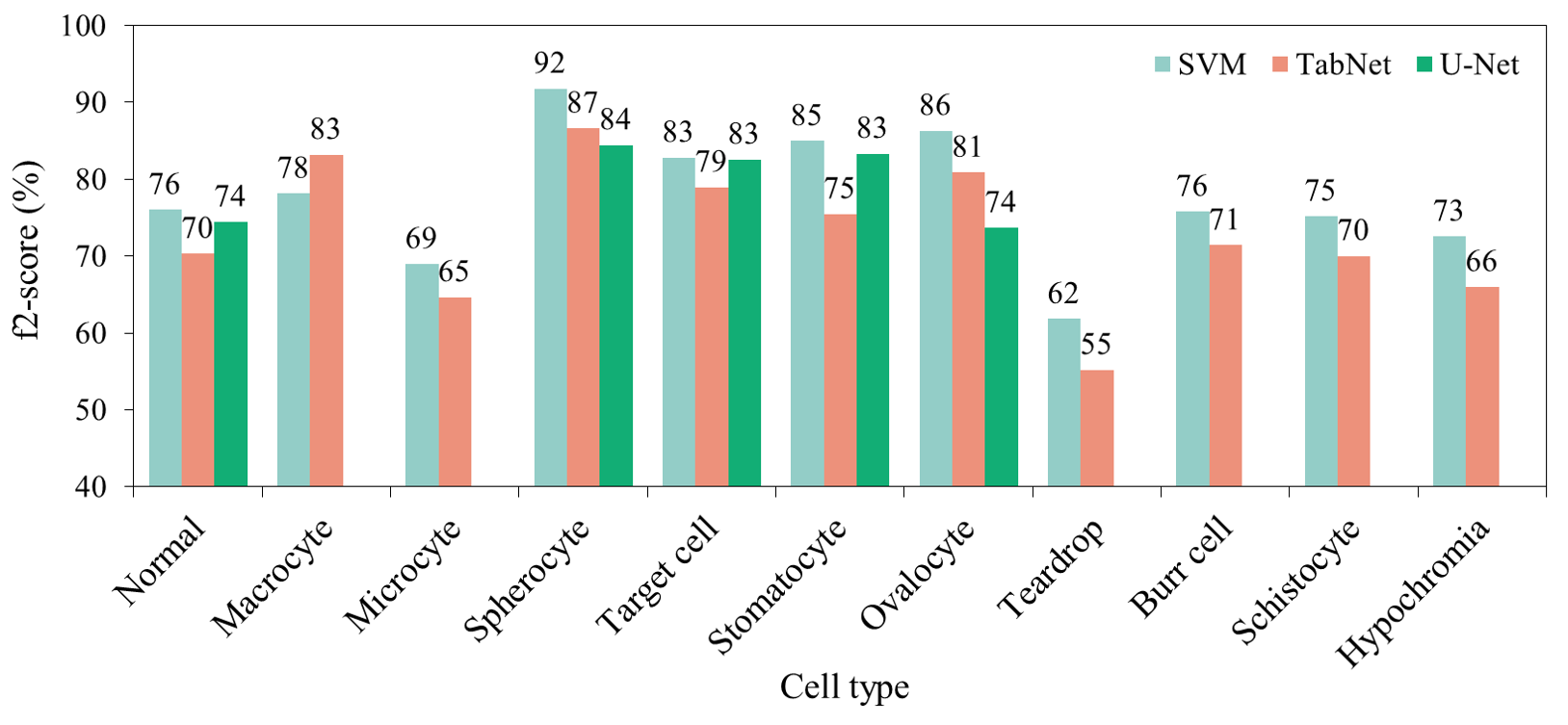}
	\caption[Comparison of multiclass models' f2-score]{Comparison of multiclass SVM, TabNet and U-Net f2-scores for individual cell types. Upsampled versions: SVM - SMOTE1 + CSL; TabNet - SMOTE1 + CSL. U-Net model stopped at epoch 20.}
	\label{fig:mc_compare_cells}
\end{figure}

The higher average f2-score of the SVM model (with SMOTE1 + cost-sensitive learning) plays out in the individual classes. Figure \ref{fig:mc_compare_cells} displays the f2-score for each individual cell type for the 11 class SVM and TabNet models, and the five class U-Net model. Except for Macrocytes, the SVM model has a higher f2-score than the TabNet model for all cell types. SVM equally tops the U-Net scores for the five cell types.

\subsubsection{Benchmarking}
\label{subsec:Benchmarking}

Table \ref{table:bench} summarises state-of-the-art with the performance metrics of the SVM, TabNet and U-Net models built in this paper. 
These different studies used different datasets and so it is not possible to compare them directly. They are displayed here to give a broad sense of where this work's models sit within the wider landscape. As well as different datasets, the cell types being classified differ across studies. This is of particular note with malaria detection studies, which can rely on the different staining the parasite causes. As discussed in section \ref{sec:dataimb}, accuracy was not found to be the most useful metric especially due to the imbalanced nature of the dataset. However many prior art studies state this metric and so it has been included here. The different dataset sizes are notable and a key limitation of some of the studies.

Whilst it is not possible to compare results directly, one can assess which metrics have been prioritised. For example \citet{Diaz2009AImages} and \citet{Devi2018MalariaSmear} in their two class SVM models appear to have aimed for higher specificity over sensitivity, whereas the opposite was true for this work. Sensitivities for the two class SVM and TabNet models are at the higher end of the range of two class SVM prior art, though with much lower specificity.
The two class U-Net model sits in the range of the two class deep learning models of \citet{Tomari2014ComputerImage} and \citet{Lee2014CellImages}.
The six class U-Net metrics for sensitivity sit at the lower end of what was achieved by \citet{Xu2017AAnemia} and \citet{Durant2017VeryErythrocytes}.

\begin{table*}[t!]
\caption{Benchmark table of state of the art and this work's models.}
\footnotesize
\centering
\begin{tabular}{@{}llcccc@{}} \toprule
    \centering
Study&	Dataset size&	Number of classes&	Accuracy (\%)&	Sensitivity (\%)&	Specificity (\%) \\\midrule
\multicolumn{2}{l}{\textbf{Classical Machine learning (SVM)}} & & & &	\\	
 \citet{Das2013MachineImages}&	888 ROIs&	6&	83.5&	96.6&	88.5 \\
 \citet{Shirazi2018ExtremeClassification}	&Train: 370. Test: 10,000 ROIs&	2&	83&	88&	85 \\
 \citet{Diaz2009AImages}&	12,557 ROIs	&2	& &	94&	99.7 \\
&	670 ROIs&	3	& &	78.8&	91.2 \\
 \citet{Devi2018MalariaSmear}&	1,300 ROIs&	2&	98.4&	94.6&	98.8 \\
 & & & & & \\
\multicolumn{2}{l}{\textbf{Deep learning}} & & & &	\\	
\citet{Korranat:red:2021} & 20,875 ROIs & 12 & 90.2\\
\citet{Xu2017AAnemia}&	7,205 ROIs	&5	&89.3&	75.5 - 93.6	& \\
&	&	8&	87.5&	84.4 - 94	& \\
 \citet{Durant2017VeryErythrocytes}&	3,737 ROIs&	10&	90.6&	92.7& \\	
 \citet{Zhang2017ImageU-Net}&	128 ROIs&	2&	82.7&	&	 \\
 \citet{Tomari2014ComputerImage}&	150 ROIs&	2&	83&	76&	76 \\
 \citet{Lee2014CellImages}&	400 ROIs&	2&	88.3&	90&	87 \\
&	360 ROIs&	4&	91& &		 \\
\multicolumn{2}{l}{\textbf{Our models}} & &  & &	\\	
SVM - Original&	Train: 12,192. Test: 3,048.&	2&	89.2&	93.6&	79.4 \\
SVM - SMOTE3+CSL(2:3)&	Train: 19,158. Test: 3,048.&	2&	88.0&	96.9&	68.3 \\
SVM - SMOTE1+CSL& 	Train: 23,370. Test: 3,048.	&11	&89.2-98.8&	62.8-92.0&	96.5-99.4 \\
TabNet - Original&	Train: 8,127. Validation: 4,065. Test: 3,048.&	2&	89.0&	92.4&	81.2 \\
TabNet - SMOTE3-3+CSL(1:2)&	Train: 12,661. Validation 4,065. Test: 3,048.&	2&	86.9&	96.5&	65.6 \\
TabNet&	Train: 15,954. Validation: 4,178. Test: 3,133.&	11&	86.3-98.2&	59.8-87.7&	94.4-98.8 \\
U-Net&	6,756 images&	2&	83.8&	83.3&	84.8 \\
U-Net&	3,546 images&	6&	87.2-96.5&	70.1-83.8&	93.6-98.1 \\\bottomrule
\end{tabular}
\label{table:bench}
\end{table*}

\section{Conclusion}
\label{chap:conc}
This paper presents an application of machine learning to detect abnormal RBC and cell-type classification. The dataset used is significantly larger and higher variation than those present in literature.
Three models were built; SVM, TabNet and U-Net. To address a problem of highly imbalanced dataset, we combined SMOTE technique with cost-sensitive learning. The abnormal cells used in this work had a wide range of shapes and shade intensities, so it is reasonable to assume the same approach could be used on other cell types.

For abnormal cell detection, the SVM and TabNet models achieved f2-scores of 94.8\%  and 94.2\% respectively, being able to correctly classify 96.9\%  and 96.5\% of abnormal cells. U-Net model achieved an f2-score of 84.9\% with a precision of 92.0\% (higher than the classification models). It was able to correctly identify 83.3\% of abnormal cells. For cell-type classification, the average f2-scores for the 11 class case reached 78.2\% and 73.0\% for the SVM and TabNet models respectively. For U-Net, the six-class model achieved an f2-score of 78.2\% with precision of 82.0\%, being able to correctly classify 77.4\% of the six classes.

Determining the `best' model requires clinical input. Defining the right balance between sensitivity to abnormal cells and the precision of tests is a difficult task. With improvements in cell region identification, treatment of overlapping cells and more extensive pixel labelling, it is proposed that all three models would realise gains in performance.

\bibliographystyle{elsarticle-harv} 
\bibliography{references.bib}





\end{document}